\documentclass[letterpaper]{article}
\usepackage{uai2020}
\usepackage[margin=1in]{geometry}

\usepackage{times}
\usepackage{natbib}
\usepackage{amsfonts}
\usepackage{mathtools}
\usepackage{graphicx}
\usepackage{caption}
\usepackage{subcaption}
\usepackage{url}
\title{The Variational InfoMax Learning Objective}

\author{Vincenzo Crescimanna \\ Department of Computer Science\\
University of Stirling\\
Stirling, UK \And Bruce Graham \\ Department of Computer Science\\
University of Stirling\\
Stirling, UK} 

\begin{document}

\maketitle

\begin{abstract}
Bayesian Inference and Information Bottleneck are the two most popular objectives for neural networks, but they can be optimised only via a variational lower bound: the Variational Information Bottleneck (VIB). In this manuscript we show that the two objectives are actually equivalent to the InfoMax: maximise the information between the data and the labels. The InfoMax representation of the two objectives is not relevant only \emph{per se}, since it helps to understand the role of the network capacity, but also because it allows us to derive a variational objective, the Variational InfoMax (VIM), that maximises them directly without resorting to any lower bound. The theoretical improvement of VIM over VIB is highlighted by the computational experiments, where the model trained by VIM improves the VIB model in three different tasks: accuracy, robustness to noise and representation quality.

\end{abstract}
\section{Introduction}
Deep neural networks are a flexible family of models that easily scale to millions of parameters and data points. Due to the large number of parameters involved, training such models while avoiding the overfitting scenario is not easy. Indeed, it is well known that minimising the naive accuracy term is not a good objective, or in general, any metric that is a distance between the predicted and the real labels. In particular, as observed in  \citep{zhang2016understanding} in the case of really powerful networks (e.g. convolutional net) it is possible to train with success, a neural net with random labels. The latter scenario means that the network is no longer learning a description (\emph{representation}) of the data with the associated labels, but a function from the weights to the labels \citep{JMLR:v19:17-646}. 

In light of this empirical observation, to bound the information conveyed in the weights, many heuristic regulariser techniques were proposed: from the classic $L_1/L_2$ weights regularisation, bounding the norm of the weights, to the more recent ones such as Dropout \citep{srivastava2014dropout} and batch normalization \citep{ioffe2015batch}, bounding the entropy of the weights. Such heuristic approaches are not easy to interpret and the hyper-parameter tuning is often not trivial.

A solution to the interpretability issue is to consider a Bayesian description, and read the neural network as a model describing the distribution associating the data to the labels. Under this perspective it is possible to relate the Dropout technique to the Bayesian inference problem \citep{kingma2015variational} and describe the neural net as an information channel \citep{alemi2016deep, achille2018information}. The latter two descriptions provide two regularised objectives: the Variational Dropout (VD) \citep{kingma2015variational} aiming to learn the optimal weights, and the Variational Information Bottleneck (VIB) \citep{alemi2016deep} aiming to learn the optimal representation of the data.
Although the two tasks: optimal weights and optimal representations are intuitively related, and VD is a special case of VIB, as observed in \citep{alemi2016deep} the objective optimised to learn the optimal weights is not the one optimised to learn an optimal representation and vice-versa. 

In this manuscript, we try to address the reasons  for such counter intuitive behaviour by studying the network from an information theory perspective. In particular we consider a third definition of the optimal network: one maximising the information theory between the data and its labels, the \emph{InfoMax} (IM) principle. The IM description has a twofold relevance: theoretical and computational. From a theoretical side it allows to identify an objective regulariser as a network capacity constraint, and in particular to prove that the optimal network is learning both optimal weights and optimal representations, i.e. VD and VIB should have the same optimum. But its main advantage is computationally, since it can be optimised directly via a variational network, the same used for VD and VIB, that are optimising a lower bound of the same principle. The theoretical advantages of the introduced objective are confirmed by the experimental results, where the model trained optimising the Variational InfoMax (VIM) performs better than VIB in three different tasks: accuracy, network robustness and representation quality.

\section{Background and related work}
 Given a dataset $\mathcal{D}$, containing a set of $N$ observations of tuples $(x,y)$, samples of the random variables $(X,Y) \sim p(X, Y)$, the goal is to learn a model with parameter $\theta\sim p(\theta)$ of the conditional probability $p(y|x, \theta)$, such that for any $x\sim p(x)$, $p(y, x|\theta) = p(y|x, \theta)p(x)$ coincides with the real $p(y, x)$. I.e., find a model $p(y|\cdot, \theta)$ such that for any distance $D$ the following objective is optimised: \begin{equation}
     \min_\theta D(p(y, x), p(y, x|\theta)).
 \end{equation}

The naive idea to minimise the negative log-likelihood, \begin{equation}\label{LogLikelihood}
\max_\theta \frac{1}{N} \sum_i^N -\log p(y^{i}|x^{i}, \theta),
\end{equation}
leads to model prone to overfit. 
Indeed, minimise the log-likelihood is equivalent to minimise the Kullback-Leibler divergence $D_{KL}(p(y|x)||p(y|x,\theta)) = \mathbb{E}_{p(y|x)}[\log p(y|x) - \log p(y|\theta, x)]$, that is optimised by a distribution $p(y|\theta)$ that does not depend by the input $x$. The latter phenomenon where the information about the labels come only from the weights is undesirable, and coincides with the complete overfitting. 

\subsection{Variational Dropout}

From the many regulariser techniques proposed, the most popular is the \emph{Dropout} one. Miming the biological behaviour of the real neural network, in \citep{srivastava2014dropout} it was proposed to train the artificial neural network using only some units i.e. to dropout some units during the training according to a distribution $p(\xi)\sim \mathcal{B}(\xi)$. As observed in \citep{baldi2018neuronal}, the dropout technique is a way to restrict the space of distributions $p(y|x, \theta)$ that the network can learn, i.e. the network \emph{capacity}. 

The original formulation with Bernoulli noise is not stable and not easy to train, so a relevant improvement was provided in \citep{wang2013fast}, where it was observed that introducing a multiplicative Gaussian noise, $ p(\xi) \sim \mathcal{N}(1, \alpha = (1-\xi)/\xi)$ behaves like the Bernoulli one, with the advantage of more robust and fast training. 
Moreover, as observed in \citep{kingma2015variational}, the introduction of the Gaussian noise allows to move the noise from the units to the weights. 

For the sake of clarity, we describe the phenomenon in the case where the network is a single layer with linear activation; the generalisation to the deep network follows naturally. Let us suppose $V$ is the weight matrix to learn and $A$ and $B$ respectively, the input and the output layers.
Then in the Gaussian dropout case we have that \begin{equation}\label{desc}
    B = (A \cdot \xi) V, \quad  \xi \sim \mathcal{N}(1, \alpha),
\end{equation}
that is equivalent, by the associative property of the direct multiplication to \begin{equation*}
    B = A \tilde{V}, \quad  \tilde{V} = \tilde{v}_{i,j} = v_{i,j}\xi_{i,j}.
\end{equation*}

By this description, the network $p(y|\cdot, \theta)$ can be read as a composition of two distributions: the regression $p(y|\cdot, W)$, a function of the last layer, $W$, and the weight inference $q(W| \cdot, \phi)$, described by the rest of the network weights $\phi$ and the noise $\xi$, i.e. $(W, \phi) = \theta$. Thanks to this description we can read the network trained minimising the negative log-likelihood with Gaussian dropout, as optimising the objective
\begin{equation}\label{VIn}
    \mathbb{E}_{q(W|\mathcal{D},\phi)}[-\log p(y|x, W)].
\end{equation}
Observing that \eqref{VIn} is a loose lower bound of the unfeasible to compute KL-divergence \begin{equation}\label{Bayes}
    - D_{KL}(p(y, x|\theta)|| p(y,x) )
\end{equation}
 \citep{kingma2015variational} provided an approximation of the KL divergence $D_{KL}(q(W| \mathcal{D}, \phi)||p(W))$ and then proposed to optimise the Variational Inference (VI)
\begin{equation}\label{VI}
\begin{split}
    	\mathbb{E}_{q(W|\mathcal{D},\phi)}&[-\log p(y|x, \theta)]\\
    	&+\mathbb{E}_{ q(\theta|\mathcal{D}, \phi)}[D_{KL}(q(W|\mathcal{D},\phi)||p(W))],
\end{split}
\end{equation}
which is a tight approximation of the term \eqref{Bayes}, with $p(W)$ the prior of the regression weights being supposed known.

\subsection{The Information Bottleneck}
In the section above we observed that the continuous noise can be moved from the latent units to the weights. Let us now leave the noise in the latent units. In this setting the network $p(y|\cdot, \theta)$ is the composition of two sub-nets: the decoder $p(y|z, W)$ and the encoder $q(z|\cdot, \phi)$, i.e. $p(y|x, \theta) = p(y|z, \theta)q(z|x, \phi)$ for any $x$, where the random variable $Z$ is defined according to \eqref{desc} as $Z = A \cdot \xi$. In light of this observation the VI objective \eqref{VI} can be rewritten as  
\begin{equation}\label{VD}
\begin{split}
	\mathbb{E}_{q(z|x, \phi)}& [-\log p(y|z, W)] \\ & +\mathbb{E}_{ q(z|x,\phi)}[D_{KL}(q(z|x, \phi)||p(z))].
\end{split}
\end{equation}

In this way we have moved our attention from the weights $\theta$, a huge number of parameters difficult to interpret, to the easier to describe latent variable $Z$. According to \citep{tishby2000information} it is possible to define an optimal network $X \to Z \to Y$, as the one learning a representation $Z$ that is a minimal sufficient statistic of $X$ for $Y$;  i.e. a description of the input data containing only the necessary information to distinguish a class element from another one. Formally, the minimal sufficient representation is the random variable $Z$ optimising the following objective:
\begin{equation}\label{tish}
        \min_\phi I(Z; X|\phi) \quad
        \text{ s.t. } I(Y;Z| W) = I(Y; X| \theta),
\end{equation}
where the conditional mutual information $I(B;A|W)$, defined as\begin{equation*}
    I(B;A|W) = H(B|W) - H(B| A, W),
\end{equation*} 
is a measure of the information conveyed by $A$ to $B$ in a channel defined by weights $W$, with conditional entropy $H(B| W) = \mathbb{E}_{p(b,w)}[-\log p(b|w)]$ denoting a measure of the information lost by $B$ about $W$.

The objective in \eqref{tish} is intractable, but it is possible to optimise a lower bound of its Lagrangian form: \begin{equation}\label{lag_vib}
        \max_{\theta, \phi} I(Y; Z| W) - \beta I(Z; X|\phi). 
\end{equation}
Indeed, observing that \begin{itemize}
    \item $H(Y| W) = H(Y)$ is constant,
    \item $H(Y|Z, W)\leq{E}_{q(z|x, \phi)} [-\log p(y|z, W)]$,
    \item $I(Z; X,\phi) \leq \mathbb{E}_{ q(z|x,\phi)}[D_{KL}(q(z|x, \phi)||p(z))]$,
\end{itemize}  
the following objective, the Variational Information Bottleneck (VIB), 
\begin{equation}\label{VIB}
\begin{split}
	\mathbb{E}_{q(z|x, \phi)}& [-\log p(y|z, W)] \\ & +\beta \mathbb{E}_{ q(z|x,\phi)}[D_{KL}(q(z|x, \phi)||p(z))],
\end{split}
\end{equation}
is a variational lower bound of the original IB in \eqref{tish}.
Let us observe that, since $I(Z; X|\phi) \leq I(Z; X,\phi)$ optimising \eqref{VIB} is equivalent to optimising a lower bound of \eqref{lag_vib}.

The VIB model in \eqref{VIB}, that is a generalisation of the VD \eqref{VD}, was independently derived in  \citep{achille2018information} and \citep{alemi2017fixing}, where it was observed that it is an outperforming regulariser, leading to robust learning (in agreement with the Bayes theory) and optimal representation quality (in agreement with the IB theory); but, as observed in \citep{alemi2016deep} the Lagrange hyper-parameter $\beta$ chosen to maximise the accuracy is not the same one used to learn robust weights and good quality representation. We suppose that this issue arises from the fact that VD \eqref{VD} and VIB \eqref{VIB} are optimising a lower bound of the respective objectives and that the choice of the prior $p(z)$ is arbitrary and often equal to the easy to compute unit variance Normal distribution.

\section{Capacity Constrained InfoMax}
\subsection{A third definition of optimal network}
In the previous section we described two different definitions of the optimal network, optimal Bayesian inference \eqref{Bayes} and minimal sufficient representation $Z$ \eqref{tish}. In this section we provide an information theoretic description of the first principle and we show that it is equivalent to the second one. 

\paragraph{The InfoMax} The mutual information between the variables $X$ and $Y$, is a constant of the system and it is defined as \begin{equation*}
    I(X; Y) := D_{KL}(p(X,Y)||p(X)p(Y)).
\end{equation*}
By property of the KL divergence, the mutual information can be decomposed as follow: \begin{equation}\label{eq_info}
\begin{split}
    I(X; Y) =& D_{KL}(p(X,Y)||p(X,Y|\theta)) +\\ & D_{KL}(p(X,Y|\theta)||p(X|\theta)p(Y|\theta)) +\\ & D_{KL}(p(X|\theta)p(Y|\theta)||p(X)p(Y)),
\end{split}
\end{equation}
where the third term is trivially zero for any $\theta$, and the second term is the conditional mutual information $I(Y; X|\theta)$. Noting that the first term is the inference objective \eqref{Bayes} to minimise, the latter problem can be rewritten as the following \emph{InfoMax} objective
\begin{equation}\label{IM}
   \max_\theta I(Y; X|\theta),
\end{equation} 
with optimum value $\theta^*$, that satisfies the following equality:
\begin{equation*}
    I(Y; X, \theta^*) = I(Y; X|\theta^*) = I(Y; X).
\end{equation*}
By the highlighted equivalence between the InfoMax and the Bayes inference, it is possible to show the equivalence between the Bayes Inference \eqref{Bayes} and the Information Bottleneck \eqref{tish}. In order to prove such an assertion it is enough to show that the optimal solution learnt by \eqref{IM} is minimal and sufficient. 

\paragraph{Proposition} \emph{The network learning parameters $\theta^*$ optimising the InfoMax objective \eqref{IM}, i.e. $I(Y; X|\theta^*) = I(Y; X)$, is learning, in the hidden layer, a minimal sufficient representation $Z$ of the input $X$ for the variable $Y$.}

\emph{Proof.} Let us observe that a model $p(y|x, \theta)$ optimises \eqref{IM} if $I(Y; \theta) = 0$, indeed $I(Y; X, \theta) = I(Y; X|\theta) + I(Y; \theta)$. Then, in order to prove the proposition it is enough to show that the parameter optimising \eqref{tish} satisfies $I(Y; \theta) = 0$.

A representation $Z$  of $X$ is \emph{sufficient} for $Y$, if there exists a function $\phi$ such that $Z = \phi(X)$, and 
\begin{equation}
    \label{suf0}
p(y|x) = p(y|\phi(x),W)\phi(x),
\end{equation} or equivalently, see \citep{cover2012elements} section 2.9, if it satisfies the following equality: \begin{equation}\label{suf1}
    I(Y;X) = I(Y;Z|W).
\end{equation} 
By the deterministic property of $\phi$, $I(Y; \phi) = 0$ for any sufficient statistic.  Then it remains to show that only for the minimal sufficient statistic $Z$ it holds that $I(Y; W) = 0$.

A sufficient statistic $Z$, is \emph{minimal} if the encoding information $I(Z,X)$ is minimal. Or equivalently, since $Z = \phi(X)$, and then $H(Z|X) = 0$ for any $\phi$, if the entropy $H(Z) = H(\phi(X))$ is minimal. Since by \eqref{suf0} we have that \begin{equation*}
    H(Y|X) = H(Y|Z,W) + H(Z),
\end{equation*}
we obtain that a minimal sufficient representation is associated to a maximal $H(Y|Z,W)$, or equivalently to a minimal $I(Y;Z,W)$. But, by \eqref{suf1}, and remembering that $I(Y; Z, W) = I(Y; Z|W) + I(Y; W)$, we have that only for a minimal sufficient representation $I(Y;W) = 0$. Q.E.D.

Thanks to this proposition we showed that the similarity between the variational objectives \eqref{VI} and \eqref{VIB} is not a causality but comes from the equivalence of the two theoretical objectives from which they were derived. Moreover, we showed that both the Bayes Inference \eqref{Bayes} and the IB \eqref{tish} problems are equivalent to the IM \eqref{IM}. Such a relationship allows us to derive an alternative variational objective that optimises directly the IM without resorting to any lower bound approximation, and moreover to highlight the role of network capacity and why it should be bounded.

\subsection{The channel capacity}
The direct optimisation of the InfoMax objective \eqref{IM} is unfeasible: it is necessary to rewrite it. Let us start by observing that the feasible to optimise negative log-likelihood, $\mathbb{E}[- \log p(y|x, \theta)]$, is equivalent to optimising the MI $I(Y; X, \theta)$, an upper bound of the desired conditional information $I(Y; X|\theta)$. Then it is useful to rewrite the IM \eqref{IM} in terms of $I(Y; X, \theta)$:
 \begin{equation}\label{constrain}
    \max_\theta I(Y; X, \theta), \quad \text{s.t. } I(Y; X, \theta)\leq I(Y;X). 
\end{equation}
In this new form we are asserting that the network capacity $C(\theta) = \sup_\theta I(Y; X, \theta)$, the maximum value that the mutual information can reach, has to be equal to the visible mutual information $I(Y; X)$. Indeed, without such a bound the information can achieve the value $H(X) + H(\theta)$, which is the scenario of pure overfitting. 

The capacity, as a function of the weights, is in general, unfeasible to compute, but given the observation made above on the relationship between weights and representation, in the following we try to write the capacity in terms of the representation. 

In a network of the type $X \to Z \to Y$, by the Data Processing Inequality the MI $I(Z; X| \phi)$ is an upper bound of $I(Y; X|\theta)$. Then, $C(\theta)\leq I(Z; X| \phi)$, moreover by equation \eqref{tish}, we have that for an optimal parameter $\theta^* = (W^*, \phi^*)$, the optimal capacity $I(Y; X)$ coincides with the encoding information  $I(Z; X, \phi^*) = H(Z)$, where the latter equality follows from the sufficiency property of $Z$. Then the InfoMax objective \eqref{IM} can be written as follows: \begin{equation}\label{alt_IM}
    \max_{W, \phi} I(Y; \phi(X), W), \quad \text{s.t. } H(\phi(X)) = I(X; Y).
\end{equation}
The alternative formulation \eqref{alt_IM} does not depend anymore on the parameter $\theta$, everything is defined in terms of the sub-networks, and this highlights the relationship between the network capacity and the entropy of the latent layer, underlying that the choice of the prior is fundamental in order to have a proper learning. Indeed, if a prior $p(z)$ with high variance is prone to over-fit, a prior with small variance will under-fit.

\subsection{The Variational InfoMax}

\paragraph{The choice of $Z$} 
In the analysis above we have seen that the choice of the prior is fundamental. This is in principle a real issue since the possible distributions are infinite. For this reason, before deriving the variational objective optimising the IM, we remember that in almost any case it is possible to restrict our attention to a standard Gaussian distribution.
Such an observation is the classic principle on which is based the Normalising Flow technique \citep{rezende2015variational}. The proof is divided into two steps: in the first it is shown there exists an invertible function $g$, where the objective is unchanged since the $I(Y; g(Z)) = I(Y; Z)$ and $H(g(Z)) = H(Z)$, see \citep{cover2012elements} chapter 2. The second step follows by the Inverse Function Theorem, where, as observed in \citep{kingma2016improved}, locally almost any function can be approximated by an invertible function.

Given these observations, we can assume without loss of generality that the latent entropy is distributed according to $p(z)\sim \mathcal{N}(0, \sigma^2 I)$, such that $H(Z) = I(X,Y)$.
In this way the IM can be re-written as \begin{equation}\label{var_IM}
     \max_{W, \phi} I(Y; \phi(X), W), \quad \text{s.t. } q(z|\phi) \sim N(0, \sigma^2 I),
\end{equation}
an objective that depends only by the variance of the prior and not by its shape.
\paragraph{The variational objective}
The advantage of the alternative representation of IM \eqref{var_IM}, is that it  can be optimised via the following variational method:
\begin{equation}\label{VIM}
\begin{split}
      \max_{\phi, W} \quad & \mathbb{E}_{q(z|x, \phi)}[p(y|z, W)] 
      -\beta D(q(z|\phi)||p(z)), \\ & \text{s.t.} \quad p(z) \sim \mathcal{N}(0, \sigma^2 I),
\end{split}
\end{equation} 
a Lagrangian relaxed form of the intractable variational objective
\begin{equation*}
\begin{split}
      \max_{\phi, W} \quad & \mathbb{E}_{q(z|x, \phi)}[p(y|z, W)] \text{, s.t. }  q(z|\phi) \sim \mathcal{N}(0, \sigma^2 I).
\end{split}
\end{equation*} 

$D$ is any function, measuring the distance between two distribution, e.g. the KL-divergence, and $\beta$ is the Lagrangian multiplier associated to the chosen divergence. 

\section{Experiments}
In this section we compare the behaviour of the same stochastic neural networks trained by optimising respectively the VIB and VIM objectives.
The section is divided into two parts: in the first one, considering the same setting analysed in \citep{alemi2016deep} of MNIST data and a fully-connected network, we show that the network trained with VIM outperforms the one trained with VIB, and that the optimal accuracy VIM model is the most robust to noise and with better quality representation. This is in agreement with the theory section where a maximally informative (maximal accuracy) model is the one learning the minimal sufficient representation (good quality representation) and minimising the Bayes Inference problem (robust to noise). In the second part, we consider a more challenging setting, CIFAR10 data and a convolutional network, to describe the role of the two hyper-parameters: the variance of the prior $\sigma^2$ and the Lagrange multiplier $\beta$. We observe that the choice of $\sigma$ is relevant for both the variational objectives, and has not to be neglected.

In all the experiments we consider as a metric $D$ in \eqref{VIM}, the Maximum Mean Discrepancy (MMD), an approximation of the KL divergence \citep{zhao2017infovae}, defined as: \begin{equation*}
\begin{split}
        MMD(q(z)||p(z)) & =\\  \sup_{f: || f||_{\mathcal{H}_k}\leq 1}& \mathbb{E}_p(z)[f(Z)] - \mathbb{E}_q(z)[f(Z)],
\end{split}
\end{equation*}
where $\mathcal{H}_k$ is Reproducing Kernel Hilbert Space associated to the positive definite kernel $k(z_1, z_2) = K/(K + \| z_1 - z_2 \|_2^2)\geq 0$, with $K$ the dimension of the latent space, i.e. $z \in \mathbb{R}^K$.

\subsection{MNIST setting}
The first setting that we consider to evaluate VIM is the one already considered by \citep{alemi2016deep} and \citep{pereyra2017regularizing}, where it is observed that the VIB objective outperforms the three most popular heuristic methods: Dropout, Label Smoothing \citep{szegedy2015going} and Confidence Penalty \citep{pereyra2017regularizing}.
Consistently with \citep{alemi2016deep}, we consider a network with  encoder modelled by an MLP with fully connected layers of the form $784-1024-1024-2K$, with ReLu activation, where $K$ is the dimension of the representation space,  and as a decoder a logistic regression with Softmax activation, i.e. $p(y; z, W) = \exp(y_c)/\sum_c\in C \exp(y_c)$, where $y = (y_c)_1^{C = 10} = Wz + b$.
Since the goal of this manuscript is not to provide the state of the art performance, nor to assert that VIM is the best regulariser in any setting, but simply to observe that VIM  is a tighter approximation of the IB objective than VIB, in all the experiments we consider the same (network) hyper-parameters used in \citep{alemi2016deep}, and we use the Adam optimiser \citep{kingma2014adam} with learning rate $10^{-4}$.

\paragraph{Accuracy} The first task of a neural network is to predict the correct label, so the first metric that we consider to evaluate the  objective is the test accuracy of the trained network.

\begin{figure}
    \centering
    \begin{subfigure}{\linewidth}
    \includegraphics[width = .85\linewidth]{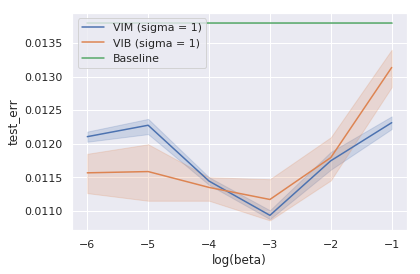}
    \caption{Comparison test-error of the same network trained with VIM and VIB as a function of $\beta$, for a fixed $\sigma = 1$}
    \end{subfigure}
    \begin{subfigure}{\linewidth}
    \includegraphics[width = .85\linewidth]{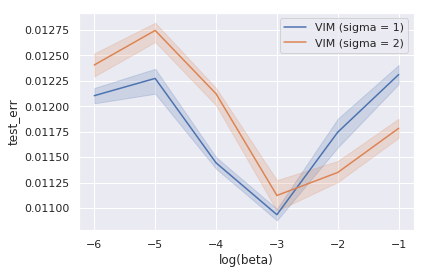}
    \caption{Comparison test-error of the same network trained with VIM with two different priors,  $\sigma \in \{ 1, 2\} $ as a function of $\beta$}
    \end{subfigure}
    \caption{Test-error on MNIST dataset for VIM and VIB trained networks}
    \label{MNIST}
\end{figure}

\begin{figure}
\begin{subfigure}{0.48\linewidth}
\centering
    \includegraphics[width = \linewidth]{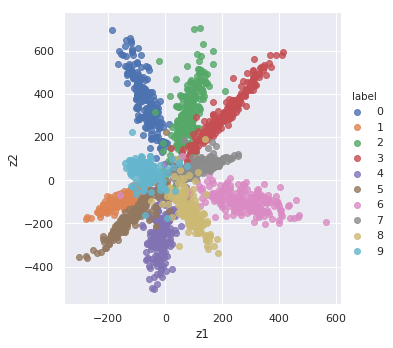}
    \caption{baseline, $adjR$: 0.81, Hoyer: 0.305 test error: 4.87\%}
\end{subfigure}
\begin{subfigure}{0.48\linewidth}
\centering
    \includegraphics[width = \linewidth]{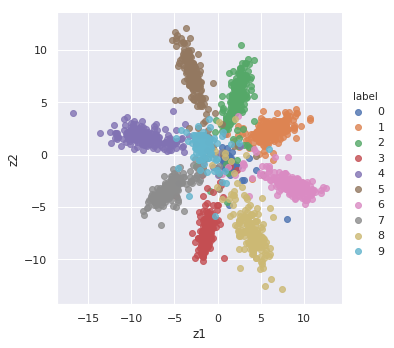}
    \caption{VIB, $adjR$: 0.81, Hoyer: 0.305, test error: 3.61\%}
\end{subfigure}
\begin{subfigure}{\linewidth}
\centering
    \includegraphics[width = .48\linewidth]{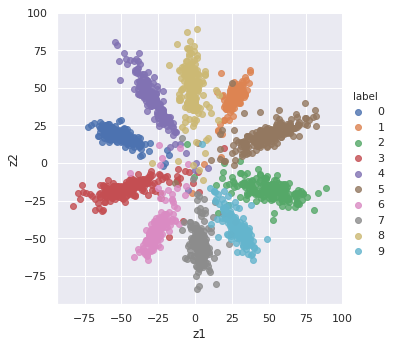}
    \caption{VIM, $adjR$: 0.901, Hoyer : 0.328, test error: 3.05\%}
\end{subfigure}
    \caption{2d learnt representation of the MNIST data, the network trained with VIM (c) is the most informative (smaller test error) and it is learning the best representation (higher Rand and Hoyer indices ) }
    \label{2d_MNIST}
\end{figure}

As we see from table \eqref{Acc} and figure \ref{MNIST} the network trained with VIM and having standard deviation $\sigma = 1$, and Lagrangian $\beta = 10^{-3}$, slightly outperforms the best VIB solution, with the same objective hyper-parameters $\beta$ and $\sigma$. Obviously, as we can see in figure \ref{MNIST} the accuracy performance is a function of both the objective hyper-parameters $\beta$ and $\sigma$, and it is  simply a coincidence that both VIM and VIB are optimised by the same couple $(\beta, \sigma)$. Indeed, as we will see in the 2d MNIST case (see figure \ref{2d_MNIST} and in the CIFAR10 case, see figure \ref{CIFAR}) the optimal hyper-parameters for the two objectives are not necessarily the same.
\begin{table}
  \centering
\caption{Comparison test-error on MNIST (smaller is better), with $Z \in \mathcal{N}(0, I)$, $I \in \mathbb{R}^{K\times K}$, $K = 256$}
  \label{Acc}
  \begin{tabular}{rc}
    {Model}&     {error (\%)} \\
    \hline
    Baseline &  1.38 \\
    Dropout \citep{alemi2016deep} & 1.34   \\
    Label Smoothing \citep{pereyra2017regularizing}   & 1.23 \\
    Confidence Penalty \citep{pereyra2017regularizing} & 1.17 \\
    VIB ($\beta = 10^{-3}, \sigma = 1$) \citep{alemi2016deep} & 1.13\\
    \textbf{VIM}($\beta = 10^{-3}, \sigma = 1$) & \textbf{1.10}    \\
        \hline

  \end{tabular}
\end{table}

\paragraph{Robustness to noise}
We observed that an optimal network is the one learning some weights that do not share any information with the label, which means that an optimal network should be robust to noise. In particular, as observed in \citep{szegedy2015going} small perturbations on the input, sometimes just a single pixel, can lead to a wrong classification. For this reason, in agreement with \citep{alemi2016deep}, we decided to measure the robustness of the network with the magnitude of corruption adversary that leads to a misclassification. 

Formally, given a network $M$ and an input $x$ with label  $C_i$ such that $M(x) = C_i$, the successful adversary $A(x)$ of $x \in C_i$,  of a (targeted) attach with target $C_j$ with $i \neq j$ is the closest element $x'$, with respect to a prescribed measure, such that $M(A(x)) = C_j$.  Defining $x'$ as the successful adversary of $x$, the robustness of the network is defined as the average distance $||x-x'||_n$, with $n \in \{1, 2, \infty\}$,  

In particular, in our experiments, consistent with the choice made in \citep{alemi2016deep}, we compute the adversary attach of the first ten zero digits in the test set with adversary target the label one, i.e. $M(A(x)) = C_1$ with $x \in C_0$, using the adversary method proposed in \citep{carlini2017towards} optimised according the $L_2$ distance.
\begin{table}
  \centering
\caption{ Distance between original and adversarial sample}
  \label{Robust}
  \begin{tabular}{rccc}
    {Model}&     {$L_2$} & $L_1$ & $L_\infty$ \\
    \hline
    Baseline &  2.20 & 713 & 0.37 \\
    
    VIB ($\beta = 10^{-3}, \sigma = 1$) &  3.38 & \textbf{752} & 0.59 \\
    VIB ($\beta = 10^{-4}, \sigma = 1$) &  3.58 & 697 &  0.63\\
    \textbf{VIM}($\beta = 10^{-3}, \sigma = 1$) & \textbf{3.70} & 700 &  \textbf{0.65}\\
    \hline

  \end{tabular}
\end{table}

As we see from the results listed in table \ref{Robust} the VIM model with $\sigma =1$ and $\beta = 10^{-3}$, obtaining best accuracy performance, is the most robust with respect to all the metrics considered but $L_1$, that, as observed in \citep{alemi2016deep} decreases when the $L_2$ distance increases.

This result gives visible evidence of the theoretical equivalence between InfoMax, the objective associated to the network accuracy, and the Bayes Inference, the objective ensuring the network robustness. Indeed, the VIM optimised network is the one having maximal accuracy and also is the most robust to adversarial attack.

\paragraph{Quality of the representation}
According to IB theory an optimal network should learn a good quality representation. The debate on what is a good representation is open; in this manuscript we follow the definition given in \citep{mathieu2018disentangling} and we consider a good representation as the one that \emph{decomposes} the hidden factors of the data. To evaluate the decomposition of the representation we consider two properties: \emph{clustering} and \emph{sparseness}. Indeed, a representation with high clustering is the one that is able to separate out the hidden factors of the visible data that allows us to recognise if an element belongs in a certain class; and a sparse representation can be thought of as one where each embedding has a significant proportion of its dimensions off, i.e. close to 0 \citep{olshausen1996emergence}. 
To evaluate the clustering, we evaluate the adjusted Rand index $adjR$ between the sets $C_i$, individuated by a classic K-means trained with 10 clusters and the set of representations associated by labels $L_i$; defining $a_i = C_i \cap L_i$ the $adjR$-index is defined as \begin{equation*}
    adjR = \frac{\sum_i a_i}{\sum C_i} \in [0,1],
\end{equation*}
yielding 1 for a complete overlapping between the clusters and the correct set and 0 if no point lies in the intersection between the two sets. 
For the sparseness we consider the Hoyer extrinsic metric \citep{hurley2009comparing}, \begin{equation*}
    Hoyer(z) = \frac{\sqrt{d} - \|z\|_1/\|z\|_2}{\sqrt{d} - 1} \in [0,1],
\end{equation*}
yielding 0 for a fully dense vector and 1 for a fully sparse vector.
Since in our experiments we are considering latent representation with different variance and high sparseness can be simply associated to a large variance, following the same approach in \citep{mathieu2018disentangling}, we evaluate the Hoyer metric on a normalised representation vector, i.e. $Hoyer = H(z/\sigma)$.

\begin{table}
  \centering
\caption{ Adjusted Rand and Hoyer index of the learned representation (higher is better)}
  \label{sparse}
  \begin{tabular}{rcc}
    {Model}&     adjR & Hoyer  \\
    \hline
    Baseline &   0.938 & 0.37 \\
    
    VIB ($\beta = 10^{-3}, \sigma = 1$) &  0.948 & 0.31 \\
    VIB ($\beta = 10^{-4}, \sigma = 1$) &  0.951 &  0.33\\
    \textbf{VIM}($\beta = 10^{-3}, \sigma = 1$) & \textbf{0.954} &  \textbf{0.41}\\
    \hline

  \end{tabular}
\end{table}
In total agreement with the accuracy and robustness performance discussed above, we see in table \ref{sparse} that the VIM trained network is the one learning the best quality representation, confirming empirically that the Bayesian Inference \eqref{Bayes}, the Information Bottleneck \eqref{tish} and the InfoMax \eqref{IM} are actually the same objective. Moreover, we observe that also in this case the VIB trained network with  $\beta = 10^{-4}$ is learning a better representation than the counterpart with $\beta = 10^{-3}$ that has optimal accuracy, suggesting that the VIB trained model cannot be optimised to be the best in all the three tasks at the same time.

\paragraph{2d latent} For the sake of completeness we considered also the case $K = 2$. This scenario is useful to visualise what the network is learning, and see the behaviour in a more challenging scenario than the one considered above. We see in figure \ref{2d_MNIST} that, as confirmed by the Hoyer and Rand indices, the learnt representations of VIM are well clustered and the intersection between the different clusters is minimal (high Rand score), and symmetric around the origin, i.e. representation close to zero and then more sparse. We conclude that, in agreement to what was observed for the case $K = 256$, a better representation corresponds with a smaller test error. Let us notice that in this case, the optimal $\sigma$ parameter differs for the two variational objectives. Such a phenomenon will be visible also in another challenging case, the CIFAR10 that we discuss below.
\subsection{CIFAR setting}
Classifying the MNIST data, although a classic benchmark, is a quite simple task, and the differences between the two variational objectives is small. The aim of this section is to show that the differences between the two considered objectives is apparent in a more challenging context, and that the choice of the variational hyper-parameters is fundamental in order to have good performance. 
For this reason in this section we decided to train a convolutional neural network to classify the CIFAR10 dataset. We take into consideration this setting since, as observed in \citep{zhang2016understanding}, a classical CNN without regulariser is prone to overfit and moreover, in \cite{achille2018information} was observed that considering the  VIB objective, the overfitting phenomenon essentially disappears and the accuracy performance is improved drastically. 
We performed the experiments considering an encoder network of four convolutional layers with filter of size $4 \times 4$ and increasing kernel size, followed with a Batch Normalization, as illustrated in table \ref{conv}, and the decoder a classic logistic as in the MNIST setting.  The structure of the network is similar to the one considered in \citep{zhang2016understanding}, and as already observed in \citep{achille2018information} the batch normalization is added only to have more stable computation, without really affecting the final results. The network is trained using Adam with learning rate starting from $10^{-3}$ and decreasing after 30 epochs by a factor of 2.

\begin{figure}
    \centering
    \begin{subfigure}{\linewidth}
    \includegraphics[width = .9\linewidth]{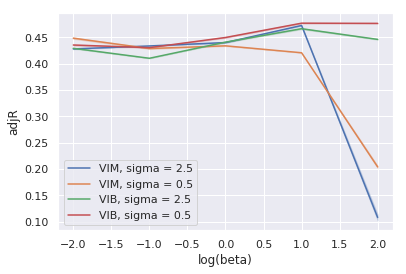}
    \caption{adjusted Rand index, for VIM and VIB, as a function of $\log(\beta)$. As expected VIM is less robust to a change of $\sigma$}
    \end{subfigure}
     \begin{subfigure}{\linewidth}
    \includegraphics[width = .9\linewidth]{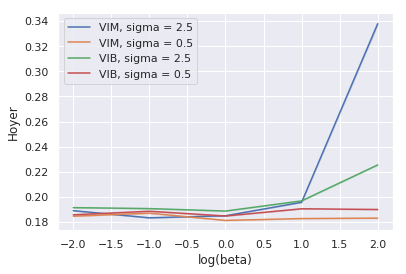}
    \caption{Hoyer index, for VIM and VIB, as a function of $\log(\beta)$. The two models have really similar results: apart the isolated case $\beta = 10^2$, the best results are obtained by the two models with highest $\sigma$.}
    \end{subfigure}
    \caption{Evaluation of the learned representation by the VIM and VIB optimisers. Note that VIM with $\sigma = 2.5$ (blue line) almost always improves over the other VIM with smaller $\sigma$; such behaviour does not hold true in the VIB case.}
    \label{CifarRep}
\end{figure}

\begin{figure}
\begin{subfigure}{\linewidth}
\centering
    \includegraphics[width = .9\linewidth]{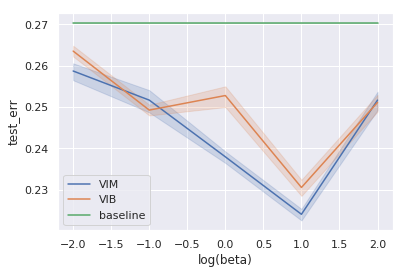}
    \caption{comparison test error, as a function of $\beta$ for a fixed $\sigma$, VIM $\sigma = 2.5$ and VIB $\sigma = 0.5$ CIFAR10}
\end{subfigure}
\begin{subfigure}{\linewidth}
\centering
    \includegraphics[width = .9\linewidth]{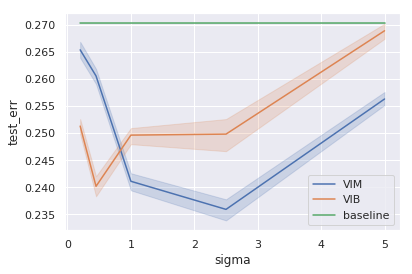}
    \caption{comparison test error as a function of $\sigma$ for a fixed $\beta = 1$, CIFAR10}
\end{subfigure}

    \caption{Comparison test error of the CNN trained with VIM and VIB, as a function of the two hyper-parameters $\sigma$ and $\beta$. We observe that a correct choice of the parameters is fundamental, but in general VIM outperforms VIB.}
    \label{CIFAR}
\end{figure}

As we can see in figure \ref{CIFAR}, the difference between the VIB and VIM trained modesl, in this scenario, is clear. Both the models are optimised by a Lagrangian parameter $\beta = 10^{-3}$, but if the VIM model has its minimum for $\sigma = 2.5$, the VIB is minimised by $\sigma = 0.5$. According with what was seen in the 2d MNIST setting, and in agreement to what was observed in the theory section: when VIB performs well, VIM cannot improve too much, instead the performance gap is larger in the more challenging case where VIB obtains results that are far from optimal. 
We conclude, by describing the quality of the learned representation, to better understand the role of the two hyper-parameters and the odd behaviour of the VIB, where the hyper-parameters associated to the best accuracy are not the same associated to the best quality representation. 
As we see in figure \ref{CifarRep}, the two objectives are learning representations of similar quality, apart from the strange behaviour of the VIM trained model for $\beta = 10^2$, that is learning really sparse representation which is then difficult to clusterise. From the results in figure \ref{CifarRep} it is possible to make two observations: the choice of the prior entropy is relevant for both the variational objectives, indeed if the VIB model is more robust, the difference in performance between the two VIB variants ($\sigma = 0.5$, $\sigma = 2.5$) is not negligible; see also the accuracy performance in figure \ref{CIFAR}. Secondly, we underline that, as observed in the MNIST framework, if in the VIM case the single model that has best accuracy is also the one learning the best representation, in the VIB context this assertion does not hold true. Indeed, if the minimal test error is obtained for $\sigma = 0.5$, see figure \ref{CIFAR}, the best Hoyer metric is achieved by the VIB with $\sigma = 2.5$  figure \ref{CifarRep}. This phenomenon is presumably a symptom of a non-accurate objective. 

\section{Conclusion}
In this manuscript we presented the Variational InfoMax (VIM), a variational objective that is optimising the InfoMax, an objective equivalent to Bayes Inference and the Information Bottleneck, maximising the information between the input data and the output labels. Differently from the Variational Information Bottleneck (VIB), that is optimising a lower bound of the IM, the VIM optimises directly the learned principle. The theoretical differences appear clear in the computational experiments, where the VIM trained models outperform the VIB trained ones, in test accuracy, network robustness and representation quality.
Moreover, the VIM derivation discloses the role of the latent prior, and in particular of its entropy that coincides with the network capacity, and then with the maximal information that can be transmitted via the network. Such observations, confirmed in the experiments, suggests to consider the variance of the prior as an hyper-parameter of the objective. In future work we will try to overcome such an issue, trying to consider the latent variance an objective term to optimise, in a fashion similar to the variational tempering technique \citep{mandt2016variational}.

The equivalence between Bayes inference and InfoMax and its easy optimisation, suggests to describe the LifeLong learning problem, learning more than one task with the same network, from an information theoretic perspective. 
In particular, in future work we will investigate a natural extension of the InfoMax to the LifeLong scenario, the \emph{conditional InfoMax}: given a network already trained for a task $A$,  and learned representation $Z_A$, train the same net for a task $B$ optimising the $Z_A$-conditioned mutual information between the visible data $x_B$ and the label $y_B$ (of task $B$), $I(Y_B, X_B|Z_A)$.
\begin{table}
    \caption{CNN architecture of the encoder network used for the CIFAR experiments}
    \centering
    \begin{tabular}{c}
    Input ($32\times 32 \times 3$)\\
    \hline
           Conv($4\times4$, 128)\\ BN $+$  ReLu\\
    \hline        
            Conv($4\times4$, 256\\ BN $+$  ReLu\\
    \hline
            Conv($4\times4$, 512)\\ BN $+$  ReLu\\
    \hline 
            Conv($4\times4$, 1024)\\
            BN $+$  ReLu\\
    \hline 
    Fully connected 
    $2K$, $K= 64$\\
    \hline
    \end{tabular}

    \label{conv}
\end{table}

\newpage

\bibliography{sample}
\newpage
\appendix

\end{document}